\documentclass{article}

\usepackage{microtype}
\usepackage{graphicx}
\usepackage{subfigure}
\usepackage{booktabs} 

\usepackage{hyperref}

\usepackage[accepted]{icml2024}

\usepackage{amsmath}
\usepackage{amssymb}
\usepackage{mathtools}
\usepackage{amsthm}

\usepackage[capitalize,noabbrev,nameinlink]{cleveref}

\theoremstyle{plain}

\theoremstyle{definition}

\theoremstyle{remark}

\usepackage[textsize=tiny]{todonotes}

\icmltitlerunning{Emergence of In-Context Reinforcement Learning from Noise Distillation}

\begin{document}

\twocolumn[
\icmltitle{Emergence of In-Context Reinforcement Learning from Noise Distillation}



\icmlsetsymbol{attink}{*}

\begin{icmlauthorlist}
\icmlauthor{Ilya Zisman}{airi,skol,attink}
\icmlauthor{Vladislav Kurenkov}{airi,inno,attink}
\icmlauthor{Alexander Nikulin}{airi,mipt,attink}
\icmlauthor{Viacheslav Sinii}{tinkoff,inno}
\icmlauthor{Sergey Kolesnikov}{tinkoff}
\end{icmlauthorlist}

\icmlaffiliation{tinkoff}{Tinkoff, Moscow, Russia}
\icmlaffiliation{airi}{AIRI, Moscow, Russia}
\icmlaffiliation{mipt}{MIPT, Moscow, Russia}
\icmlaffiliation{inno}{Innopolis University, Kazan, Russia}
\icmlaffiliation{skol}{Skoltech, Moscow, Russia}

\icmlcorrespondingauthor{Ilya Zisman}{i.zisman@airi.net}

\icmlkeywords{reinforcement learning, in-context learning, Machine Learning, ICML}

\vskip 0.3in
]



\printAffiliationsAndNotice{\icmlEqualContribution} 

\begin{abstract}
Recently, extensive studies in Reinforcement Learning have been carried out on the ability of transformers to adapt in-context to various environments and tasks. Current in-context RL methods are limited by their strict requirements for data, which needs to be generated by RL agents or labeled with actions from an optimal policy. In order to address this prevalent problem, we propose AD$^\varepsilon$, a new data acquisition approach that enables in-context Reinforcement Learning from noise-induced curriculum. We show that it is viable to construct a synthetic noise injection curriculum which helps to obtain learning histories. Moreover, we experimentally demonstrate that it is possible to alleviate the need for generation using optimal policies, with in-context RL still able to outperform the best suboptimal policy in a learning dataset by a 2x margin. 

\end{abstract}

\section{Introduction}

Reinforcement Learning (RL) has achieved great success in recent years, from showing superhuman performance in games \cite{fuchs2021super, schrittwieser2020mastering} to successfully learning complex real-world tasks \cite{brohan2022rt1, herzog2023deep}. At the same time, RL has a persistent problem of severe sample inefficiency. For example, in order to learn to match human performance in StarCraft 2, an agent may need to process tens of billions of frames for a single game \cite{vinyals2019grandmaster}. At the same time, a trained agent cannot generalize to new tasks, so mastering another game requires the entire process to be repeated from scratch. In order to enable fast and reliable generalization and adaptation in RL agents, Meta-RL was introduced. This framework is designed to teach agents the process of learning itself, enabling them to learn how to learn to accomplish unseen tasks \cite{finn2017model, duan2016rl, wang2016learning}. Recent approaches leverage specially constructed datasets that enable learning the task in-context by interacting and adapting to the environment \cite{laskin2022context, lee2023supervised, kirsch2023towards, sinii2023context}. These approaches are referred to as in-context RL.

\begin{figure}[t!]
    \includegraphics[width=\columnwidth]{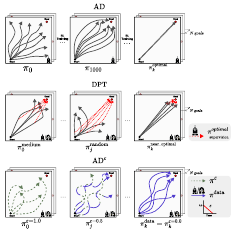}
    \caption{Data acquisition for in-context RL training. While other in-context RL methods either train thousands of single-task RL algorithms to obtain their learning histories (AD) or pretrain on optimal actions (DPT), our approach AD$^\varepsilon$ alleviates these problems introducing synthetic noise injection curriculum by which learning histories are generated. Algorithms trained on this kind of data can not only generalize to unseen tasks, but also outperform the best available policy in data ($\pi^\text{data}$).}
    \label{fig:main}
\end{figure}

It is important to note that the construction of curated data that can enable in-context RL may be considered troublesome. \citet{laskin2022context} show that the data must reflect the process of learning a task, rather than simply presenting a sequence of optimal actions. One might view this as the data containing different policies, which are ordered from the least to most effective. To make this possible, authors train numerous (1000+) single-task RL agents to obtain their learning histories. This poses a challenge in terms of computational and time constraints, one which can make the task unfeasible for some problems. At the same time, \citet{lee2023supervised} claim that in-context RL can be made possible by pretraining on random datasets. However, they set a requirement for obtaining the optimal policy for the task, which, in essence, also requires training a multitude of RL agents. If it were possible to alleviate those problems by generating learning histories or leveraging suboptimal trajectories that are commonly available, it would make training in-context RL algorithms significantly easier.



In our work, we present \textbf{AD$^\varepsilon$}, a data collection strategy inspired by \citet{brown2020better} that produces learning histories for any policy by adding noise to it. In doing so, we ensure that each successive transition in data illustrates a slightly improved policy than the previous one, resembling the learning process. Our approach allows in-context RL to not only learn from the expert policies, but also demonstrate near-optimal performance from suboptimal policies. Similarly to \citet{laskin2022context} and \citet{lee2023supervised}, we tested our approach on grid-world and 3D environments and achieved more than \textbf{2x} improvement when compared to the best policies available in data.


As our core contribution, we show that distilling actual learning algorithms or possessing the optimal policy is not required to enable in-context RL. Instead, it is enough to provide in-context RL agents with a simulated learning process, eliminating the need for training numerous RL agents. We summarize our main findings below.



\begin{figure*}[t!]
\vskip 0.2in
\begin{center}
    \centerline{\includegraphics[width=\textwidth]{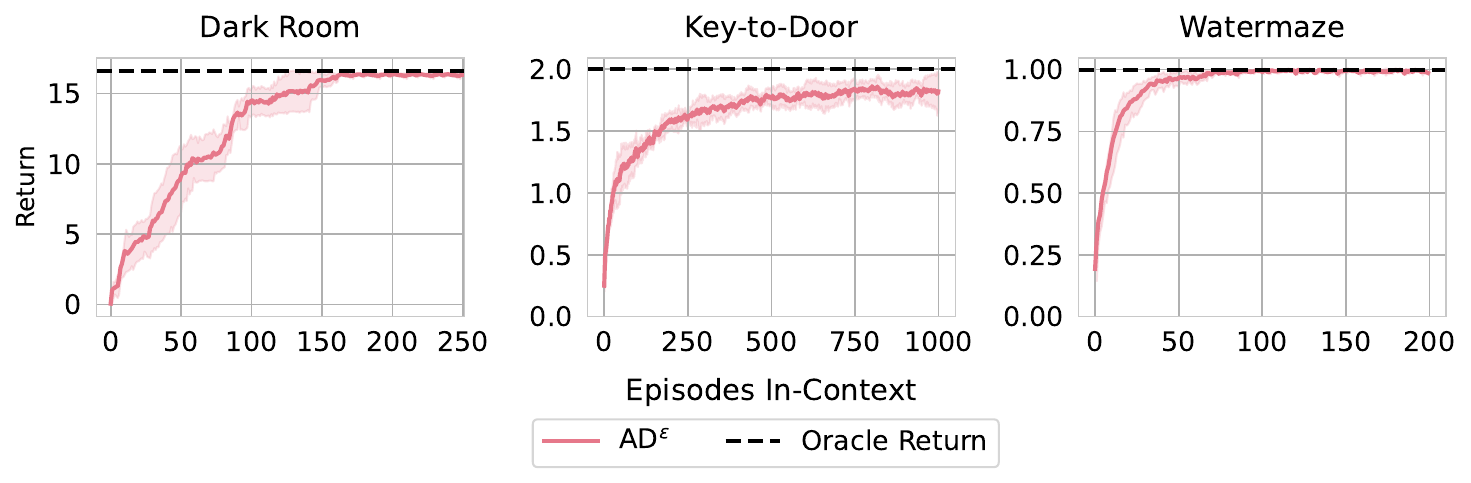}}
    \caption{The performance of AD$^\varepsilon$ on test environments. The agent must find unseen goals by memorizing visited states and rewards. The main difference from the standard approach used in in-context RL is that we generate learning histories by infusing noise, therefore eliminating the need of training thousands of single-task RL agents. Here, we demonstrate that our data collection strategy is able to provide suitable data for training in-context RL models. The mean performance of data generating oracles is shown for comparison. The performance is averaged across three seeds with the shaded regions of one std.}
    \label{fig:optimal_results}
\end{center}
\vskip -0.2in
\end{figure*}

\begin{itemize}
    \item \textbf{Distilling noise-induced trajectories of a demonstrator gives rise to in-context reinforcement learning.}
    We discuss the limitations of current approaches and propose a new method of data acquisition that enables in-context RL in \Cref{sec:backgrnd}.
    \item \textbf{Supervised pre-training does not require trajectories from RL agents or supervision with optimal actions for emergence of in-context RL.} We show that it is possible to replace actual learning histories or eliminate the need for the optimal policy without negatively impacting the performance of in-context RL algorithms. We illustrate our findings in \Cref{fig:optimal_results}.
    \item \textbf{It is possible to improve upon suboptimal demonstrators with pure in-context learning after supervised pre-training.}
    We examine the ability of in-context RL to not only emerge from suboptimal trajectories, but to significantly outperform them. The results are shown in \Cref{fig:subopt_results}.
\end{itemize}



\section{What Enables In-Context Reinforcement Learning?}
\label{sec:backgrnd}
Generally, the training of an in-context RL model consists of two main parts. The first one is the pre-training of a Transformer in a standard supervised learning fashion to predict the next token in a sequence. The second is a data collection strategy that actually enables in-context RL. There are two main approaches to data collection, which are discussed in the following paragraphs.




\paragraph{Data Collection with RL Algorithms.}
\citet{laskin2022context} have proposed training in-context RL on learning histories of RL algorithms. The authors claim that in order for in-context RL to successfully emerge, having only expert demonstrations is insufficient. To enable in-context RL, the authors supply a Transformer with the histories of how single-task RL agents learn a task, and then distill the policy improvement operator that resides between subsequent transitions in these learning histories. After pre-training on learning histories, a Transformer is able to explore the environment and generalize to unseen tasks entirely in-context without explicitly updating its weights. The authors call their method Algorithm Distillation (AD).

More formally, if we assume that a dataset $\mathcal{D}$ consists of \textit{learning histories}, then

$$
\mathcal{D}: = \Bigl\{ \left( \tau_1^g, ..., \tau_n^g \right) \sim \left[ \mathcal{A}^{source}_g | g \in \mathcal{G} \right] \Bigl\},
$$

where $\tau_i^g = \left( o_1, a_1, r_1, ..., o_T, a_T, r_T  \right)$ is a trajectory generated by a source algorithm from $\mathcal{A}^{source}_g$ \ for a goal $g$ from a set of all possible goals $\mathcal{G}$, and $o_i, a_i, r_i$ are observations, actions and rewards respectively.



It is important to note that the above approach makes it necessary to train thousands of single-task RL algorithms in order to successfully obtain suitable pre-training data. This data acquisition process can prove challenging, since RL algorithms are prone to instability in learning and are sensitive to hyperparameters.

\paragraph{Data Collection with Optimal Actions.}
The approach of \citet{lee2023supervised} puts forward another idea, called Decision Pretrained Transformer (DPT). Rather than pre-training a Transformer on a mixture of expert and non-expert data, the authors suggest training the model solely on optimal actions. This method populates the context with various environment interactions, including random ones. The authors provide proof demonstrating that pre-training a Transformer in this manner enables it to perform a Posterior Sampling (PS), an effective Bayesian algorithm that has been considered computationally infeasible \cite{osband2013more}. This algorithm is viewed as a generalization of Thompson Sampling for MDPs. 

However, despite the above approach being sound, it necessitates high-quality training data. For certain tasks where the environment is sufficiently complex, constructing the optimal policy is intractable. There are often numerous demonstrations available from suboptimal actors, but such data cannot be used to pre-train DPT. The authors mention a potential workaround that involves training single-task RL agents, but it brings us back to the previously mentioned challenges associated with large-scale RL training.

In summary, both AD and DPT data acquisition methods enable in-context RL, but each approach comes with its own considerations. To achieve success with these methods, it is necessary to either carefully train a multitude of RL algorithms or, even more challenging, to obtain the optimal policy. With this in mind, one might pose the following question: Is it possible to democratize data acquisition? In our paper, we demonstrate that generation of learning histories is not only feasible, but that this data can be created using suboptimal policies. Moreover, an in-context RL algorithm pre-trained on suboptimal learning histories is capable of outperforming even the best available policy in the dataset. 

\section{AD$^\varepsilon$}
\label{sec:ad_eps}


When observing the trajectories in a learning history buffer, one can notice that  $\mathcal{D}$ are subject to a certain order $\left[R(\tau_i) < R(\tau_j) \ | \ i < j \right],$ given there are enough parameter updates between $\tau_i$ and $\tau_j$, $R(\tau) = \sum_{i=0}^{T}{r_i}$. As a result, we propose constructing a new dataset, $\mathcal{D}^{\varepsilon}$, by infusing random noise into the given policy, thus generating new trajectories that resemble learning histories. Specifically, we run a policy in an environment where, at each step, it takes a random action with a probability of $\varepsilon$, or it adheres to its action with a probability of $1 - \varepsilon$. A critical aspect to note is that we schedule $\varepsilon$ in such a way that, at the onset of the trajectory, the demonstrator executes only random actions ($\varepsilon_\textrm{start} = 1$), while at the end, it performs actions without randomness ($\varepsilon_\textrm{end} = 0$). Our approach is hugely inspired by \citet{brown2020better}, where they employ similar methods of policy ordering in the Imitation Learning setting. However, we do not learn any reward models or RL algorithms explicitly, leaving the Transformer to do in-context policy distillation.

To test our approach, we take Algorithm Distillation \cite{laskin2022context} approach as the base that learns from the data produced by synthetic noise injecting curriculum. Throughout our work, we refer to it as AD${^\varepsilon}$.

\section{Experimental Setup}
\subsection{Model}
We use the GPT-2 causal architecture \cite{radford2019language} from the CORL package \cite{tarasov2022corl} as the backbone. For the input, we merge multiple episodes together to form a multi-episodic memory. We use states, actions, rewards triplets as the model input. Note that we cannot put full multi-episodic sequences into memory, since Transformers have quadratic requirements for sequence length. Instead, we subsample sequences into smaller parts. We also concatenate the embeddings of states, actions and rewards, inflating sequence length by a factor of 3. To avoid confusion, we report sequence length before the concatenation of embeddings. As the order of transitions matters, we apply positional encoding to the input sequences. The exact hyperparameters of the model can be found in \Cref{apndx:model_hyper}.

\subsection{Generation of Learning Histories}
\label{subsec:gen}
\paragraph{Optimal Policy.} As previously discussed in \Cref{sec:ad_eps}, we infuse random noise to the given policy. Before starting, we set $\varepsilon=1.0$, as if the agent is not familiar with the environment and acts completely on random. After that, we start generating data by running agents in the environment with different goals. At each time step, we generate a random number $\omega$ from a uniform distribution, $\omega \sim \mathcal{U}_{[0, 1]}$. Then, if $\omega < \varepsilon$, the agent takes a random action from the action space that is taken to the environment. Otherwise, it executes an action according to its policy. After the action selection phase, we decrease $\varepsilon$ according to its schedule and then repeat the process until enough trajectories are collected. Note that we decrease $\varepsilon$ after each action, not once per fully completed episode. We collect the last 10\% of the data with $\varepsilon=0$ (the best available policy), as it improves the stability of the training process.

\paragraph{Suboptimal Policy.}
We employ a similar approach in order to simulate a suboptimal policy. The key distinction is that, while $\varepsilon$ is scheduled to decrease from 1 to 0 in the optimal scenario, in case of suboptimal policy, we do not reduce $\varepsilon$ entirely to zero. This ensures that even the best policy available in the data still contains a predefined amount of noise, and is therefore not optimal. The exact values of $\varepsilon$ that correspond to a certain level of performance differ for each environment and are discussed separately in the next subsection.

\paragraph{Learning Pace.}
The proposed data collection strategy makes it possible to have a high degree of control over the learning pace, i.e., how fast $\varepsilon$ decays through the trajectory. \citet{shi2023crossepisodic} showed that the granularity of the task-difficulty learning curriculum can affect the in-context learning ability. We speculate that regulating the performance schedule in learning histories is also crucial, a factor yet to be fully explored in previous studies. This may be due to the fact that exerting fine control when training RL agents is a complex challenge. Regardless, this feature has a direct impact on the learning abilities of in-context agents. 

We compute the decay rate by regulating how many histories (full trajectories until termination) are generated for a single goal. The exact number of histories is reported in \Cref{apndx:data_hyper}.

\subsection{Environments}
\label{sec:env}
To demonstrate in-context learning abilities, we choose environments that cannot be solved zero-shot after pretraining \cite{laskin2022context, lee2023supervised}. This can only be achieved in environments with many tasks where it is possible to specify a set of training goals, leaving the rest for the test time. An additional consideration is that the episode length must be of an appropriate size so that a multi-episodic training sequence can be formed.


\textbf{Dark Room.} 2D POMDP with discrete state and action spaces \cite{laskin2022context}. The grid size is $9 \times 9$, where an agent has 5 possible actions: up, down, left, right and do nothing. The goal is to find a target cell, the location of which is not known to the agent in advance. The episode length is fixed at 20 time steps, after which the agent is reset to the middle of the grid. The reward $r=1$ is given for every time step the agent is on the goal grid, otherwise $r=0$. The agent does not know the position of the goal, hence it is driven to explore the grid. In total, there are 81 goals, of which we use 65 for training and 16 for evaluation.


\textbf{Key-to-Door.} Similar to \textit{Dark Room}, but it first requires an agent to find an invisible key and then the door. Without a key, the door will not open. The reward is given when the key is found ($r = 1$) and once the door is opened (also $r = 1$), after which the game terminates. The agent then resets to a random grid. The maximum episode length is 40, and since we can control the location of the key and door, there are around 6.5k possible tasks.

\begin{figure}[h]
    \centerline{\includegraphics[width=0.95\columnwidth]{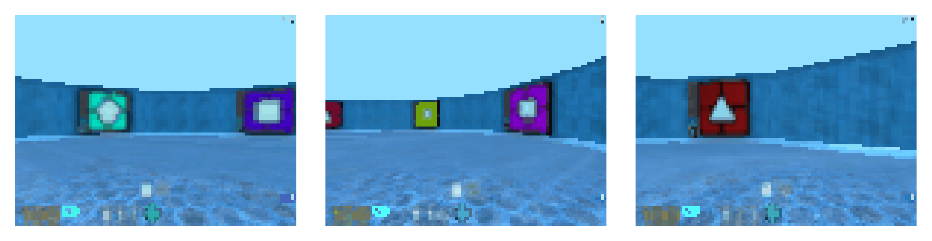}}
    \caption{Examples of Watermaze input images.}
    \label{fig:obs_water}
\end{figure}

\textbf{Watermaze.} 3D POMDP with a continuous state space and a discrete action space. An agent is placed in a playground similar to DarkRoom, where it searches for an invisible platform \cite{morris1981spatial}. When found, the platform rises and the agent receives $r = 1$, then the agent is reset to the middle. The episode length is 50, and the observation size is $3 \times 72 \times 96$. To navigate, the agent uses the walls of the playground, painted in different colors. There are a total of 8 available actions. The goal space is continuous. More details on implementation of this environment can be found in \Cref{appndx:watermaze}.

\begin{figure*}[t]
\vskip 0.2in
\begin{center}
    \centerline{\includegraphics[width=\textwidth]{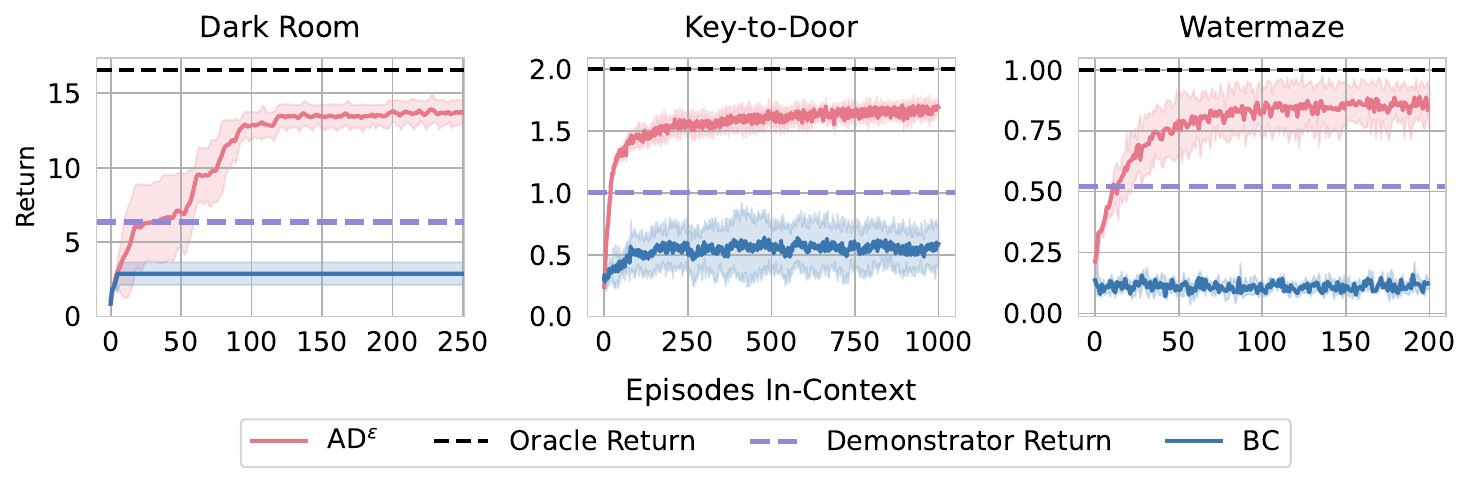}}
    \caption{The performance of AD$^\varepsilon$ pretrained on the data generated by suboptimal policies. We show that in-context agents can outperform even the best policy available in the data by a large margin, which highlights the ability of in-context RL agents to learn without pretraining on optimal actions. In these experiments, we use a generating policy that is 40\% (for Dark Room) and 50\% (for Dark Key-to-Door and Watermaze) of the optimal performance in the environment. Its mean performance is shown in a blue dashed line. We observe a substantial improvement when compared to the suboptimal policy: +120\% (6.36 → 14.08) in Dark Room, +74\% (1.0 → 1.74) in Dark Key-to-Door, +76\% (0.52 → 0.92) in Watermaze. For further comparison, we show Behavior Cloning that is unable to generalize to unseen tasks. The mean performance of the best policy available in the data is shown in light blue. The performance is averaged across 3 seeds with the shaded regions of one std.}
    \label{fig:subopt_results}
\end{center}
\vskip -0.2in
\end{figure*}

\subsection{Evaluation}
After pre-training, we evaluate the in-context reinforcement learning ability of the agent. The evaluation process resembles the standard in-weight learning, with the key difference being that the learning itself happens during evaluation. When interacting with the environment, the agent populates the Transformer's context with the latest observations. At the start of the evaluation, the context is empty. Note that the agent's context is cross-episodic, which makes it possible for the agent to access transitions in previous episodes. We report the cumulative reward that the agent achieved during fixed in-context episodes. We employ $\mathcal{G}_{\textrm{train}}, \mathcal{G}_{\textrm{eval}}, \mathcal{G}_{\textrm{test}}$ task split, $\mathcal{G}_{\textrm{train}}, \mathcal{G}_{\textrm{eval}}$ during the pre-training phase to select the best model, $\mathcal{G}_{\textrm{test}}$ during evaluation. For evaluation, we take 200 tasks for Dark Key-to-Door and Watermaze. For Dark Room, there are only 16 unseen tasks left, so we do not split them into test and eval and simply take the last checkpoint for reporting. We evaluate Dark Room for 250 episodes, Dark Key-to-Door for 1000 episodes, and Watermaze for 200 episodes. Average performance is calculated across 3 seeds with the shaded regions of one standard deviation. 

\section{Experiments}
\label{sec:exp}
In our experiments, we first show that it is possible to distill the generalized optimal policy from noise, thus making it possible to replace learning histories from RL agents with generated ones. Then we move on to the experiments with suboptimal policies, showing that in-context reinforcement learning is able to outperform them even without the optimal policy. We also test our method on a pixel-based domain and show that it is still able to sustain its abilities despite the more challenging environment. Lastly, we perform an analysis of two factors, (1) what the minimum policy performance should be to enable in-context RL, and (2) the influence of learning pace in data on the final in-context learning performance.

\paragraph{Can in-context reinforcement learning emerge from noise-generated trajectories?}
\label{para:optimal_exp}
To generate noise-induced trajectories, we need to obtain the generating policies. For Dark Room and Dark Key-to-Door, we use oracles that know the goal location and navigate there. We run these policies in the environment, then add noise as described in \Cref{subsec:gen}.

\Cref{fig:optimal_results} shows the emergence of in-context generalization from noise. For illustrative purposes, we also present the mean performance of the optimal policies on the same tasks in the environment. The in-context agent can generalize to unseen goals both in Dark Room and Key-to-Door. This emphasizes that the learning histories that are generated from noise can successfully acquire essential properties required for in-context reinforcement learning. Thus, it is indeed possible to enable in-context RL from noise distillation.

\paragraph{Is in-context reinforcement learning capable of leveraging suboptimal trajectories and outperforming them?}
The demonstrations can often be suboptimal, same as the policies extracted from them. Is it possible to learn an in-context algorithm if we run AD$^\varepsilon$ on suboptimal policies? To show that it is possible, we first obtain suboptimal policies by employing slightly different approaches from \Cref{subsec:gen}. Unlike with optimal policies, where we schedule $\varepsilon$ from $\varepsilon_\textrm{start} = 1$ to $\varepsilon_\textrm{end} = 0$ and thus arrive at an ideal policy, here, we deliberately set $\varepsilon_\textrm{end} \neq 0$. This makes it so that there is noise remaining in the data, and therefore the behavior of the policy in the data cannot be considered optimal. We determine the value of $\varepsilon_\textrm{end}$ by examining the maximum return that the policy can achieve relative to the optimal policy. The exact values can be found in \Cref{apndx:data_hyper}.

In \Cref{fig:subopt_results}, the in-context learner significantly outperforms the best policy available in the data. In the Dark Room environment, the performance increase is \textbf{+120\%} (6.36 → 14.08), and for Key-to-Door, we observe a +74\% gain (1.0 → 1.74). Interestingly, in the Key-to-Door environment, a suboptimal demonstration policy rarely visits a door with a key. However, the in-context agent still manages to learn how to explore the grid and find a door. 

For comparison purposes, we included the mean return achieved by the generating policies as well as by the Behavior Cloning algorithm.

\paragraph{Is it possible for an in-context agent to learn from noise in pixel-based environments?}
To verify the effectiveness of our method, we tested it in Watermaze, a complex 3D environment with a continuous task set. This environment presents a considerable challenge due to the navigational difficulties caused by the first-person view that significantly limits the agent's field of view. Furthermore, the agent is required to learn to explore a continuous space rather than a grid. To obtain the policies for Watermaze, we train PPO-LSTM agents from stable-baselines3 \cite{stable-baselines3}. The rest of the data collection procedure remains the same for optimal and suboptimal experiments.

In \Cref{fig:optimal_results}, we show that the in-context agent can successfully be trained on noise-induced data from images. Moreover, when the generating policies are suboptimal, there is a performance boost of +76\% (0.52 → 0.92), meaning that the agent can successfully improve upon suboptimal policies and is able to learn a multi-task policy. We illustrate the results of training on suboptimal data in \Cref{fig:subopt_results}.

\begin{figure*}[t]
\vskip 0.2in
\begin{center}
    \centerline{\includegraphics[width=0.9\textwidth]{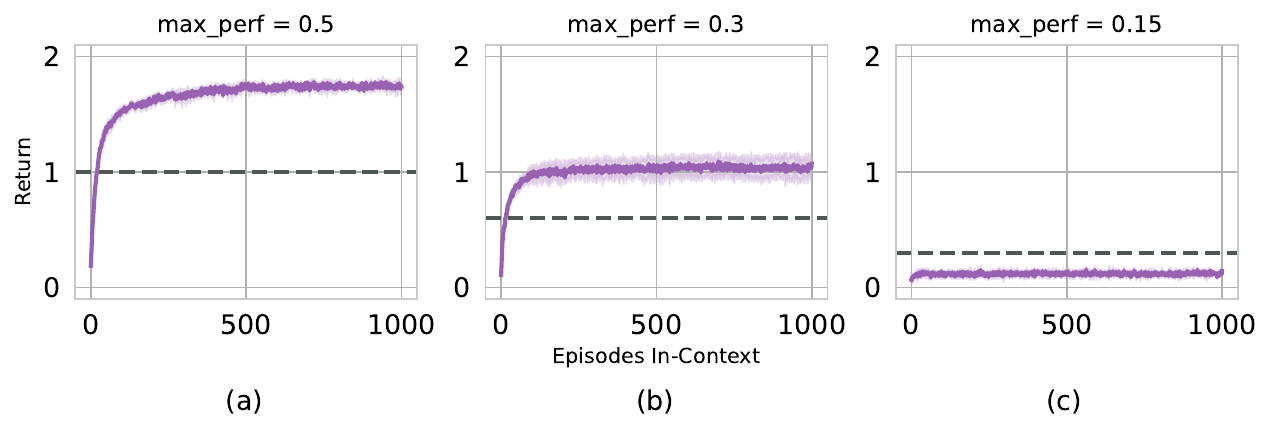}}
    \caption{The performance of AD$^\varepsilon$ for different suboptimal generating policies in the Key-to-Door environment with the following two subgoals: find a key, open a door using the key. To test the limits of in-context agents, we generated three datasets, each representing different performance relative to the maximum reward. This was made possible by scheduling $\varepsilon$ to a non-zero number, so that the mean performance of generating policies is bound by $\textrm{max\_perf}$. As one can observe in (b), the in-context agent manages to outperform even a significantly suboptimal policy used for data generation. However, it is important to point out that, in order for the in-context agent to successfully learn and achieve both subgoals (locating a key and door), the data must contain sufficient examples of both tasks. In the case of (b), the mean reward of the data-generating trajectories is 0.6, indicating that the agent generating the data rarely encounters a door. As a result, the in-context learner also struggles to learn the task of finding the door. Similarly, in (c), the in-context agent fails to learn effectively due to the same lack of diverse examples in the training data. We plot the mean returns of generating policies in a greenish-gray dashed line. The AD$^\varepsilon$ performance is averaged across three seeds $\pm$ 1 std.}
    \label{fig:anal_eps}
\end{center}
\vskip -0.2in
\end{figure*}

\paragraph{To what degree can a policy be suboptimal, and still allow in-context reinforcement learning to emerge?}
To identify the extent of suboptimality that can still enable in-context RL, we fix the total amount of data, goals and the pace of learning to be the same, allowing variation only in the performance of the final policy. Since it is possible to regulate the suboptimality by tweaking $\varepsilon_\textrm{end}$, we obtained three policies with 50\%, 30\% and 15\% from an optimal policy's maximum performance. We chose Key-to-Door as an environment because of its rich structure of subgoals. Namely, an agent needs to find several grids, rather than learning to find a single spot.

\Cref{fig:anal_eps} illustrates the varying performance of the generating policy in the Key-to-Door environment, testing the limits of the in-context agent to generalize on non-perfect data. The average return for generating policies is shown in dashed lines. One can observe that the further the policy deviates from optimality, the poorer the performance of the in-context agent becomes. We hypothesize that in order for the learning of a generalized policy to be effective, it is crucial for the dataset to include a significant number of instances where each subgoal is achieved. Consider the case with $\textrm{max\_perf} = 0.15$, where the agent has learned only to collect the key, but fails to learn the task of finding the door. The issue here may be that there are not enough examples of the agent actually reaching the door, given that the mean policy return is 0.3. This could mean that the in-context agent did not have sufficient data to learn how to successfully reach the door. It is worth noting that, even with severely suboptimal policies ($\textrm{max\_perf} = 0.3$), the in-context ability still arises, and the agent shows an improvement when compared to the suboptimal policy.

\paragraph{How does the learning pace in data affect the in-context learning ability?} 
To identify a pattern, we set the number of data samples and goals to be the same as in the best suboptimal runs with $\textrm{max\_perf} = 0.5$ in the Key-to-Door environment. We modify the decay rate of $\varepsilon$ by changing the number of full trajectories (i.e., transitions until termination) that are collected per goal. As a result, we arrive at a different learning pace in the generated data.

\begin{figure}[t!]
    \centerline{\includegraphics[width=0.8\columnwidth]{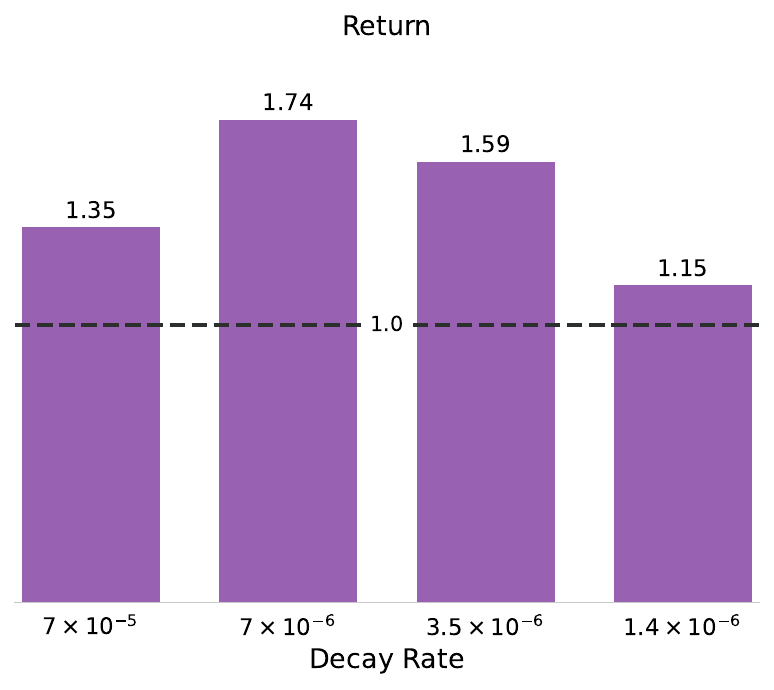}}
    \caption{AD$^\varepsilon$ with different decay values of $\varepsilon$ during data generation. The performance of a generating policy is fixed at 50\% from the optimal policy in the Key-to-Door environment. The decay rate is controlled by changing the number of full episodes for a single goal. From our results, one can observe a clear pattern where the learning pace has an ideal value around a specific decay rate. When the decay rate is too fast, there are not enough explorative actions in the data, since the trajectories start to match the best available policy too soon. However, the slower the decay, the more obfuscated the policy improvement becomes, making the in-context agent unable to fully capture it and thus resulting in worse performance. We report the mean returns across three seeds, for each of the decay values. The generating policy mean return is shown by a dashed line.}
    \label{fig:anal_hist}
\end{figure}

\Cref{fig:anal_hist} shows that the learning pace has an ideal value. With a fast schedule, where $\varepsilon$ decays with the rate of only $7 \times 10^{-5}$ (200 episodes), the in-context agent is still able to learn and generalize, outperforming the generating policy. However, the margin of improvement increases and peaks with a decay of $7 \times 10^{-6}$ (2000 episodes), after which the performance improvement starts decaying again. This leads us to assume that there may be not enough exploration within the data at a fast decay rate, since it quickly converges to the generating policy without showing much variability. The opposite is the case for slow decay rates ($3.5 \times 10^{-6}; 1.4 \times 10^{-6}$ (4k and 10k histories, respectively). If we slow down the improvements in data too much, this may obscure them, making distillation significantly more difficult. It is also important to consider the limitations of sequence length size, as improvement might not be identifiable at slower decay rates if the sequence length is short.




\section{Related Work}

\paragraph{Transformers in Reinforcement Learning.}

Our work draws heavily on the application of the Transformer architecture \cite{vaswani2017attention} in Reinforcement Learning. As \citet{agarwal2023transformers} state in their survey, Transformers are an increasingly popular approach to solving RL tasks, including world models \cite{iris2023}, offline RL \cite{yamagata2022qlearning}, and building general multi-task and multi-modal models \cite{reed2022generalist}. In this paper, we take a close look at the latter, exploring the capabilities of Transformers for one-shot generalization. The rise of Transformers in RL and decision-making can be attributed to the works of \citet{chen2021decision} and \citet{janner2021sequence}. The Trajectory Transformer focuses primarily on learning the dynamics of the environment, while the Decision Transformer formulates the problem as next-token prediction based on the previous interactions with the environment. Although both models are suitable for single-task environments, they cannot generalize to unseen tasks. Later research addressed the problem of multitasking and proposed improvements to these methods. \citet{lee2022multi} suggested pretraining Transformers on Atari games, and later fine-tuning them for unseen games to achieve near-human performance. Another study by \citet{lin2022switch} proposed training various components within a single Transformer to predict rewards-to-go as well as the next states and actions when given observations and task identifiers. However, both approaches still require fine-tuning for unseen tasks and have not demonstrated one-shot generalization capabilities. The pursuit of creating a generalist multi-task RL agent capable of adapting and learning while interacting with the environment has led to the emergence of in-context RL, which is closely related to Meta-Reinforcement Learning.

\paragraph{Meta-Reinforcement Learning.}

Meta-Reinforcement Learning is a family of RL methods aimed at learning generalized and sample-efficient algorithms on top of inefficient and single-task common RL agents. A trained meta-RL model is one that has learned how to learn. In other words, this model is able to explore an environment and can adapt to new goals. Gradient-based approaches make this possible by training a meta-agent using gradients of many on-policy RL algorithms \cite{finn2017model}. Memory-based approaches leverage the hidden structure of RNNs \cite{hochreiter2001learning, duan2016rl, wang2016learning} to train a generalized algorithm using online data produced by multiple RL algorithms interacting within the environment, each with a different goal. 

\paragraph{In-Context Reinforcement Learning.}
In-context RL is another approach to building generalist agents. Recent works use the in-context learning ability of Transformers \cite{von2023transformers, dai2022can, wei2022emergent} that has to do with Bayesian inference \cite{muller2021transformers, xie2021explanation}. It should be noted that in-context RL research places particular emphasis on data manipulation. \citet{laskin2022context} enables in-context learning by deploying RL agents in various environments to record their learning histories, which are then used to distill the policy improvement operator. In contrast, \citet{lee2023supervised} concentrates on pre-training a Transformer with optimal actions, using the surrounding context to infer the dynamics of the environment. Unlike traditional meta-RL, these approaches focus on in-context learning. Instead of updating the model weights for new tasks by fine-tuning them, these methods allow the Transformer to perform internal gradient descent, enabling zero-shot generalization to unseen tasks.

At the same time, in-context learning has its limitations, with the most notable one being the substantial data requirements. In order to efficiently generalize on a set of tasks, it needs large and diverse datasets of learning histories that consist of different tasks for downstream RL algorithms to be trained on. On top of that, the data must contain a sequential improvement from one transition to the other in order to enable the model's ability to explore. To address the above problems, researchers have proposed an approach that involves reusing existing trajectories by ranking them according to inherent metrics of quality, specifically rewards \cite{liu2023emergent}. This method has the potential to produce an order similar to the one in learning histories. However, a flaw in the evaluation protocol employed by the authors was found \cite{zisman2023agentic}, leaving the efficacy of this method in reutilizing existing demonstrations an unresolved question. Other methods exist that aim to address the problems of in-context learning by generating a few learning histories and augmenting them with random projections to achieve diverse data for in-context pretraining \cite{kirsch2023towards, kirsch2022generalpurpose}. While our work follows a similar approach, we create learning histories instead of working with pre-existing ones. The key contribution of AD$^\varepsilon$ is its ability to address the challenges of in-context learning by generating appropriate data without needing to acquire learning histories of a multitude RL agents.

The ability of in-context RL to distill different types of algorithms \cite{wang2024transformers} opens up new perspectives on its applicability, namely, learning near-optimal policies from suboptimal data. This problem has been vastly studied in Imitation Learning setup, but yet to be fully discovered for in-context RL. The approach we have our inspiration from, proposed by \citet{brown2020better}, offers to inject noise to the suboptimal policy extracted from data in order to obtain an ordering from the worst policy to the best. Then, a reward function is learned from this synthetic ordering that is later used for a standard RL training. In our approach, we do not explicitly learn neither reward function nor downstream RL agents, rather we employ the noise injection curriculum to construct learning histories. There also exists a theoretical result for in-context learning discovered by \citet{jeon2024informationtheoretic}, which states that the misspecification error of Bayesian predictors trained on optimal and suboptimal data approaches zero, provided that the number of tasks and the sequence length tend toward infinity. In our paper, we provide indirect corroboration of this implication from a parallel research field, demonstrating that it is indeed possible for in-context RL agents to learn a policy that outperforms the demonstrator.

\section{Conclusion}
In this work, we addressed the limitations of current in-context RL research, namely, the problems faced by large-scale training of single-task RL agents \cite{laskin2022context}, and having to obtain the optimal policy for efficient in-context pretraining \cite{lee2023supervised}. Through experiments using our approach AD$^\varepsilon$, we illustrate that it is possible to enable in-context RL using a synthetic noise injection curriculum that makes it possible to generate learning histories. We also showed that it is possible to efficiently train in-context agents even with suboptimal policies. This possibility has been predicted theoretically for in-context learning in a recent study by \citet{jeon2024informationtheoretic}. In addition, we showed that the pace of policy improvement in data directly affects the performance of an in-context RL agent. As a result, we believe that our work is a valuable step in the direction of democratizing the data requirements needed for in-context RL to emerge. 


Although we have attempted to address one of the limitations of AD and DPT, there are others that remain. It can be insightful to try in-context learning in more diverse environments, such as Crafter \cite{hafner2021benchmarking}, XLand-MiniGrid \cite{nikulin2023xlandminigrid} or NetHack Learning Environment \cite{kuttler2020nethack, hambro2022dungeons, kurenkov2023katakomba}.

\section*{Acknowledges}
We are grateful to Ilya's spouse, Margarita, for her great mental support and graphic advice.

\section*{Impact Statement}
The work presented in this paper is aimed at advancing the field of Machine Learning. There can be many potential societal consequences of our work, none of which we feel must be specifically highlighted here.

\bibliography{icml2024}
\bibliographystyle{icml2024}

\newpage
\appendix
\onecolumn

\section{Comparison with AD and ED}
We also conducted additional experiments comparing our method with the original data collection strategy from \citet{laskin2022context} for Dark Room and Key-to-Door environments. In order to cover all AD's benchmarks, the Expert Distillation (ED) is also included for comparison.

\begin{figure}[h!]
    \centerline{\includegraphics[width=0.8\columnwidth]{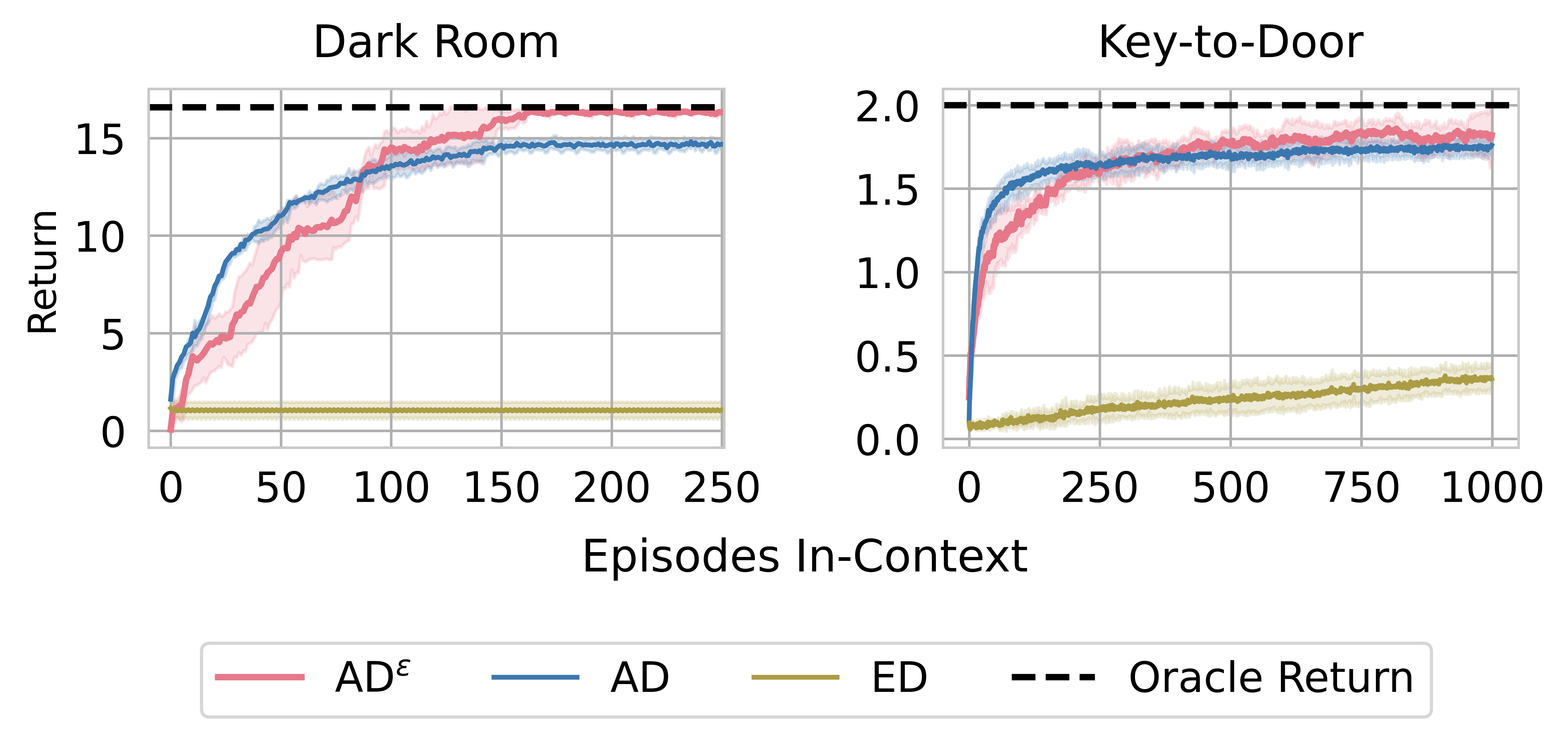}}
    \caption{AD$^\varepsilon$ acts on par with AD and significantly outperforms expert distillation (ED).}
    \label{fig:ad-baseline}
\end{figure}

As it could be seen in fig, out method enables in-context learning ability in transformer, thus acting on par with AD. At the same time, learning only from expert trajectories does not benefit for the in-context learning ability.

\section{Learning from extracted policy}
In the main text we introduce noise into the oracle policy to generate suboptimal data. In this section we conduct additional experiments when the generating policy is truly suboptimal to more closely simulate real-world policy acquisition. Instead of using the optimal policy for the noising process, we did the following: we trained a behavior cloning agent on the policy that had been injected with noise in advance, up to 50\% performance from the optimal policy. Then, we used this BC agent to produce learning histories data according to the our protocol with the epsilon curriculum. We compare this method with the one from the main text and see that both of them perform on the same level.

\begin{figure}[h!]
    \centerline{\includegraphics[width=0.4\columnwidth]{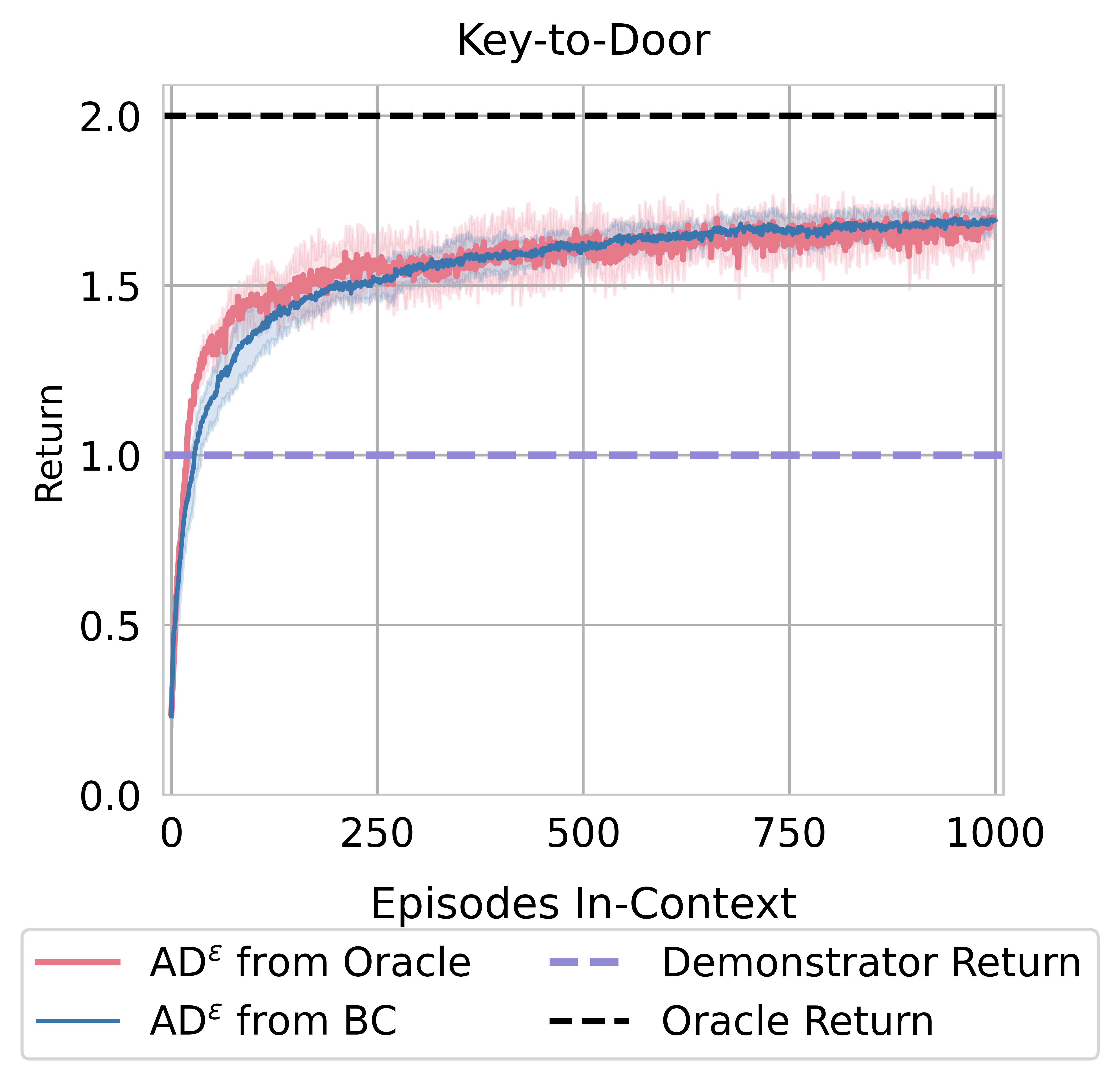}}
    \caption{AD$^\varepsilon$ can learn from a BC extracted policy as well as from the oracle.}
    \label{fig:from-bc}
\end{figure}

\section{Decreasing starting noise}
One could argue that starting with $\varepsilon_\textrm{start} = 1$ can lead agents into a collapse they cannot escape from (i.e. robots, self-driving cars). To show that our method is applicable in different settings, we do and experiment with $\varepsilon_\textrm{start} = 0.5$. Fig shows there is no difference between these two approaches.

\begin{figure}[h!]
    \centerline{\includegraphics[width=0.6\columnwidth]{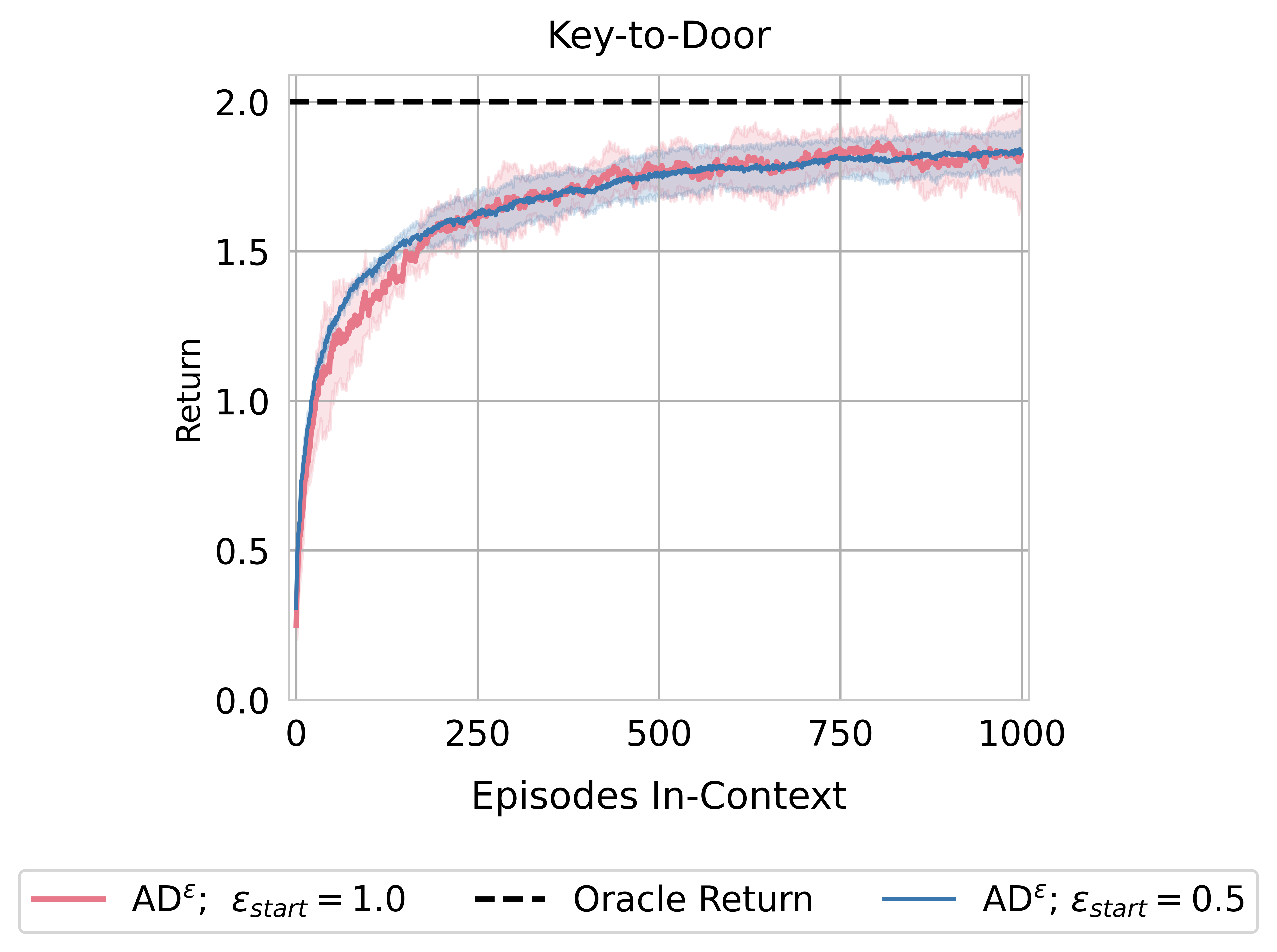}}
    \caption{Different values of $\varepsilon_\textrm{start}$ for AD$^\varepsilon$ are also suitable for data generation.}
    \label{fig:start-eps}
\end{figure}

\section{More on Watermaze Setup}
\label{appndx:watermaze}
Watermaze is build upon the DMLab engine \cite{beattie2016deepmind}, we use a predefined level $\texttt{contributed\//dmlab30\//rooms\_watermaze}$. To make interaction with the environment more convenient, we used the Shimmy package \cite{jun_jet_tai_2023_8140744} that provides a gym-like API for DMLab. We set 1 FPS for rendering the environment. We discretize continuous actions in the following manner: the agent's $x$ position can be changed by going forward, backward with full speed or no-op: \texttt{[0, 1, -1]}, same for $y$ position. The angle which camera turns is set for 12 pixels, that is enough to rotate the camera around the agent in 5 moves. The action space is \texttt{gym.Discrete} with 8 actions in total, 6 single actions: forwards, backwards, left, right, camera rotation to the left, to the right. And 2 combined actions: forwards + left rotation, forwards + right rotation. Other actions, such as jump, crouch or fire are not used. Also, we exclude goals in 200-unit radius, since they can be accessed in 10 or less actions.

\section{Additional Figure for Decay Rate Analysis}

\begin{figure}[h!]
    \centerline{\includegraphics[width=0.5\columnwidth]{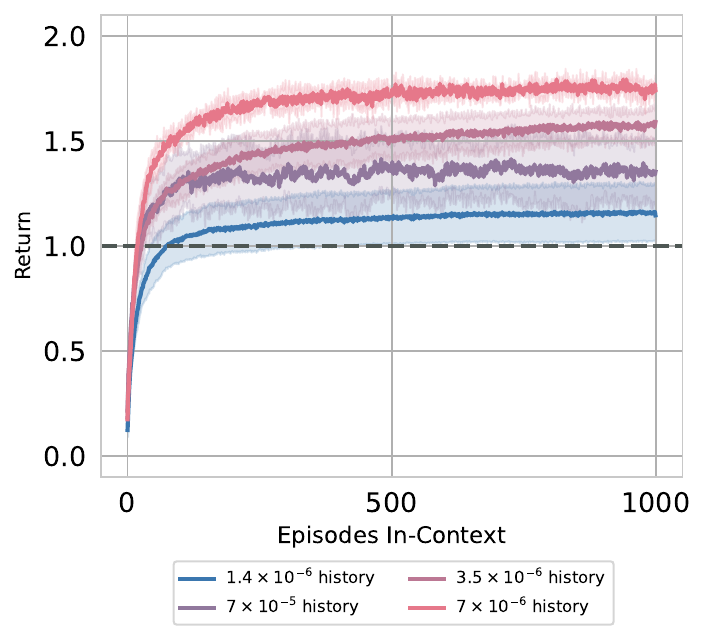}}
    \caption{In-context learning plot for different decay rates. The in-context agent is still quite stable and able to learn despite different decay rates.}
    \label{fig:anal_hist_2}
\end{figure}

\pagebreak
\section{Model Hyperparameters}
\label{apndx:model_hyper}

\begin{table}[h!]
\centering
\begin{tabular}{@{}lccc@{}}
\toprule
\multicolumn{1}{c}{\textbf{Param}} & \multicolumn{1}{l}{\textbf{Dark Room}} & \textbf{Dark Key-to-Door} & \multicolumn{1}{l}{\textbf{Watermaze}} \\ \midrule
embedding\_dim     & 64     & 64     & 64     \\
hidden\_dim        & 512    & 512    & 1024   \\
num\_layers        & 4      & 4      & 8      \\
num\_heads         & 4      & 4      & 4      \\
seq\_len           & 80     & 160    & 250    \\
attention\_dropout & 0.5    & 0.5    & 0.5    \\
residual\_dropout  & 0.1    & 0.1    & 0.1    \\
embedding\_dropout & 0.3    & 0.3    & 0.3    \\ \midrule
learning\_rate     & 3e-4   & 3e-4   & 3e-4   \\
betas                              & 0.9, 0.99                              & 0.9, 0.99                 & \multicolumn{1}{l}{0.9, 0.99}          \\
clip\_grad         & 1.0    & 1.0    & 1.0    \\
batch\_size        & 512    & 512    & 64     \\
num\_updates       & 300k   & 300k   & 200k   \\
optimizer          & \multicolumn{3}{c}{Adam} \\ \bottomrule
\end{tabular}
\caption{Transformer hyperparameters.}
\label{tab:hyp-model}
\end{table}

\section{Data Hyperparameters}
\label{apndx:data_hyper}

\begin{table}[h!]
\centering
\begin{tabular}{@{}lcccccc@{}}
\toprule
                 & \multicolumn{2}{c}{Dark Room} & \multicolumn{2}{c}{Dark Key-to-Door} & \multicolumn{2}{c}{Watermaze} \\ \toprule
Max. Performance & 1.0           & 0.5           & 1.0              & 0.5               & 1.0           & 0.5           \\ \midrule
Max Steps        & \multicolumn{2}{c}{20}        & \multicolumn{2}{c}{40}               & \multicolumn{2}{c}{50}        \\
Train Goals      & \multicolumn{2}{c}{65}        & \multicolumn{2}{c}{1000}             & \multicolumn{2}{c}{1056}      \\
\begin{tabular}[c]{@{}l@{}}Learning Histories \\ per Goal\end{tabular} & \multicolumn{2}{c}{25000} & \multicolumn{2}{c}{2000} & \multicolumn{2}{c}{500} \\
Max $\varepsilon$          & 0.0           & 0.5           & 0.0              & 0.75              & 0.0           & 0.7           \\ \bottomrule
\end{tabular}
\caption{Env and data generation hyperparameters.}
\label{tab:my-table}
\end{table}

\pagebreak
\section{Data Generation Algorithm}
\begin{algorithm}[h!]
   \caption{Data Generation}
   \label{alg:data_gen}
\begin{algorithmic}
   \STATE {\bfseries Input:} number of learning histories $\texttt{num\_hist}$, number of goals $\texttt{n\_goals}$, acting policy $\pi^{\textrm{data}}$, action space $\mathcal{A}$
   \STATE Initialize $\varepsilon = 1.0$
   \STATE Initialize $\texttt{eps\_diff} = \frac{ 1.0}{\texttt{num\_hist}}$
   \REPEAT
   \FOR{$i=1$ {\bfseries to} $\texttt{num\_hist}$}
   \REPEAT
   \STATE $\omega = \mathcal{U}_{[0, 1]}$ 
   \IF{$\omega < \varepsilon$}
   \STATE Sample an \texttt{action} uniformly from $\mathcal{A}$
   \ELSE
   \STATE Sample an \texttt{action} from $\pi^{\textrm{data}}$
   \ENDIF
   \STATE Observe \texttt{state, reward, done} by acting with \texttt{action} in the environment
   \STATE Save \texttt{state, action, reward}
   \STATE Set $\varepsilon = \max(\varepsilon - \texttt{eps\_diff}, \ 0)$
   \UNTIL $\texttt{done}$ is $true$
   \ENDFOR
   \UNTIL{all $\texttt{n\_goals}$ are collected}
\end{algorithmic}
\end{algorithm}

\end{document}